\documentclass{article}

     \usepackage[preprint,nonatbib]{neurips_2020}

\usepackage[utf8]{inputenc} %
\usepackage[T1]{fontenc}    %
\usepackage{hyperref}       %
\usepackage{url}            %
\usepackage{booktabs}       %
\usepackage{amsfonts}       %
\usepackage{nicefrac}       %
\usepackage{microtype}      %
\usepackage{graphicx}
\usepackage{tabularx}
\usepackage{epsfig}
\usepackage{subcaption}
\usepackage{amsmath}
\usepackage{amssymb}
\usepackage{grffile}
\usepackage[group-separator={,}]{siunitx}
\usepackage{algorithm,algpseudocode}
\usepackage{float}

\title{Improved Search Strategies with Application to Estimating Facial Blendshape Parameters}

\author{%
  Michael H.~Bao \\
  Department of Computer Science\\
  Stanford University\\
  Stanford, CA 94305 \\
  \texttt{mikebao@stanford.edu} \\
  \And
  David A.~B.~Hyde \\
  Department of Mathematics \\
  UCLA \\
  Los Angeles, CA 90095 \\
  \texttt{dabh@math.ucla.edu} \\
  \And
  Xinru Hua \\
  Department of Computer Science\\
  Stanford University\\
  Stanford, CA 94305 \\
  \texttt{huaxinru@stanford.edu} \\
  \And
  Ronald Fedkiw \\
  Department of Computer Science\\
  Stanford University\\
  Stanford, CA 94305 \\
  \texttt{fedkiw@cs.stanford.edu} \\
}

\begin{document}

\maketitle

\begin{abstract}
    It is well known that popular optimization techniques can lead to overfitting or even a lack of convergence altogether; thus, practitioners often utilize ad hoc regularization terms added to the energy functional.
    When carefully crafted, these regularizations can produce compelling results.
    However, regularization changes both the energy landscape and the solution to the optimization problem, which can result in underfitting.
    Surprisingly, many practitioners both add regularization and claim that their model lacks the expressivity to fit the data.
    Motivated by a geometric interpretation of the linearized search space, we propose an approach that ameliorates overfitting without the need for regularization terms that restrict the expressiveness of the underlying model.
    We illustrate the efficacy of our approach on minimization problems related to three-dimensional facial expression estimation where overfitting clouds semantic understanding and regularization may lead to underfitting that misses or misinterprets subtle expressions.
\end{abstract}

\section{Introduction}\label{sec:introduction}

There exist many standard methods for finding the solution to an unconstrained nonlinear optimization problem; one typically iteratively solves a sequence of linear problems to progress towards the solution of the fully nonlinear problem.
For example, the Gauss-Newton method \cite{madsen2004methods} aims to minimize an energy function $f(x)$ by iteratively solving a linear problem $A^T A \delta x = A^T b$, where $A$ is the Jacobian of $f$ evaluated at the current parameter state $x$, $b$ is the current residual, and $\delta x$ is the desired step in parameter values to move towards the solution.
This process is repeated until convergence.
Although this linear problem minimizes $\|A\delta x - b\|_2$, it is unclear in practice that minimizing this L2 norm is the best way to make progress towards the value of $x$ that best minimizes $f(x)$. 
In fact, the Levenberg-Marquardt and the Dogleg trust-region methods blatantly modify the least squares problem, claiming to make the Gauss-Newton method more robust \cite{lourakis2005levenberg,marquardt1963algorithm}.

Even when one can trust the linear problem to provide progress towards the solution or can modify the linear problem to be more trustworthy, one must still solve the linear problem and deal with its potential over- or under-determinedness.
For example, one may add regularizations such as the L2-norm of the parameters to ensure that $A^T A$ is nonsingular \cite{hoerl1970ridge} or add the L1-norm of the parameters to create sparsity \cite{boyd2004convex} in the spirit of minimum-norm solutions to underdetermined problems.

Furthermore, accurately minimizing the energy functional may not even lead to the desired solution since the energy is often only suggestive of intent.
This is the classical problem of overfitting, e.g.\ minimizing loss while obtaining large generalization errors on holdout data.
In the specific case of facial expression estimation, matching pre-drawn contours may minimize the energy functional at the cost of implausible deformations of the rest of the face.
Thus, domain-specific regularizers such as Laplacian smoothing of surface meshes \cite{bhat2013high} or anatomical simulation of the underlying muscle model \cite{cong2016art} help to alleviate overfitting.
In the case of training neural networks, one can utilize early stopping, dropout, and batch normalization\footnote{More recent works have instead noted that the advantage of batch normalization is to enable faster training by creating more stable gradients \cite{NIPS2018_7515}.} to achieve models that better generalize to more data \cite{goodfellow2016deep}.

To summarize, linear subproblems may not lead to the solution, often require modifications for solvability anyway, and may come from an energy functional which is only at best suggestive of intent.
Thus, we take an aggressive approach to solving these linear subproblems by first pruning away any coordinates that are uncorrelated with the right hand side as motivated by least angle regression (LARS) \cite{efron2004least}.
The other coordinates are then estimated in a coordinate descent fashion \cite{shi2016primer} eliminating the need to regularize for solvability.
We aim to find parameters that give a meaningful interpretation of observed data; thus, using uncorrelated directions to make progress on linear subproblems seems potentially unwise.
The notion that solving linear systems might not help to solve a nonlinear problem is not new; in many nonlinear problems one looks at linearizations only to motivate a strategy for tackling the nonlinear problem---a quite common example is eigensystem consideration for solving hyperbolic conservation laws, see e.g.\ \cite{toro2013riemann} and the references therein.

To the best of our knowledge, our approach is a novel search strategy for nonlinear optimization problems, with efficacy illustrated on problems in the context of facial expression estimation, especially when targeting rotoscope curves.
The end goal is to determine a set of model parameters that represent the correct semantic (human-interpretable) meanings, including pose and expression, while minimizing the excitation of spurious degrees of freedom.

\section{Related Work} \label{sec:related_work}

\textbf{Regularization:}
When working with ill-posed problems, regularization is used to prevent the model from overfitting to the data \cite{goodfellow2016deep,hastie2001elements}.
Ridge regression \cite{hoerl1970ridge} and LASSO \cite{tibshirani1996regression} are two popular methods for regularizing linear least squares problems; they add constraints equivalent to introducing L2 and L1-norm regularization respectively on the model parameters \cite{hastie2001elements}.
Using L1-norm regularization has the additional benefit of keeping the solution sparse \cite{boyd2004convex,donoho2006most}.
Additionally, regularization will help ensure that the normal equations used to solve least squares are well-conditioned.
However, regularization forces a trade-off between matching the data and having smaller parameter values.

\textbf{Coordinate Descent:}
At each iteration, coordinate descent chooses a single coordinate direction to minimize.
This allows the coordinate descent algorithm to avoid null spaces making it an attractive option for use on ill-posed, poorly conditioned problems.
The column can be chosen stochastically \cite{nesterov2012efficiency} or deterministically.
Popular deterministic methods for choosing the next search direction include cyclic coordinate descent \cite{krylov1966solution}, the Gauss-Southwell (GS) and Gauss-Southwell-Lipschitz rule \cite{nutini2015coordinate}, and the maximum block improvement (MBI) rule \cite{chen2012maximum}.
Generally, either a line search is performed to determine the step size or a fixed step size/learning rate is used.
However, the cost of a line search is prohibitive when the function takes a long time to evaluate (e.g.\ in the case of simulations) so we instead use coordinate descent to solve the linear problem found at every iteration of Gauss-Newton.
Instead of looking at a single column at a time, block coordinate descent can be used to update multiple columns simultaneously \cite{tseng2001convergence}; however, with block coordinate descent, regularization may still be needed to avoid ill-posed problems when the block of columns does not have full rank.
A more complete overview of coordinate descent methods can be found in \cite{shi2016primer,wright2015coordinate}.

\textbf{Correlation:}
Although coordinate descent algorithms can avoid the null space, they may still potentially choose many poorly correlated coordinates in place of fewer more strongly correlated coordinates.
Using correlation to choose the next coordinate to add to the model can alleviate this problem and is the central idea behind MBI \cite{chen2012maximum}, forward and backward stepwise regression \cite{derksen1992backward}, and LARS \cite{efron2004least}.
The latter statistical regression methods are often used to gain better prediction accuracy and interpretability of the model \cite{hastie2001elements}.
However, LARS converges to the least squares solution \cite{efron2004least}, which often means eventually using uncorrelated coordinates.

\textbf{Faces:}
When solving for facial blendshape parameters \cite{lewis2014practice}, L2-norm regularization of the parameters is commonly used by practitioners to obtain more reasonable solutions due to its ease of use in commodity least squares solvers \cite{blanz1999morphable,cao2014facewarehouse,li2010example,thies2016face2face}.
Others instead choose to bound the blendshape weights between some minimum and maximum value (typically $-1$ and $1$ or $0$ and $1$) \cite{bhat2013high,hsieh2015unconstrained,li2013realtime}.
The usage of the L1-norm regularization is relatively rare, even though it results in sparser solutions \cite{bouaziz2013online,ichim2015dynamic,neumann2013sparse}.
Other methods use Laplacian regularization \cite{bhat2013high,huang2011leveraging} or anatomically motivated regularization \cite{wu2016anatomically} to constrain the deformation of the face.

\begin{figure}[!ht]
\centering
\includegraphics[width=0.48\linewidth]{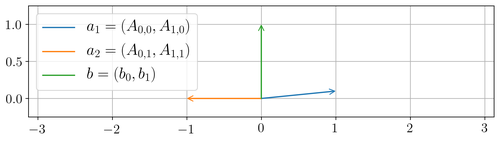}
\includegraphics[width=0.49\linewidth]{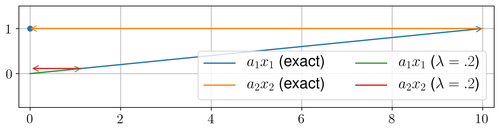}
\caption{Left: A visualization of the columns of $A$ as well as $b$ for the linear problem $Ax = b$ from Section 1 when $b = \protect\begin{bmatrix} 0 & 1\protect\end{bmatrix}^T$.
Right: The exact solution depicted by $a_1 x_1$ and $a_2 x_2$, and the (incorrect) regularized solution with $\lambda = .2$.}
\vspace*{-.2in}
\label{fig:motivation0}
\end{figure}

\section{Motivation} \label{sec:motivation}

Consider $Ax = b$ where $A = \begin{bmatrix} 1 & -1 \\ .1 & \num{1e-6} \end{bmatrix}$ and $b = \begin{bmatrix} 0.0 & 1.0 \end{bmatrix}^T$.  The two columns of $A$ get overdialed by the exact solution of $x_1 = 1 / (\num{1e-6} + .1)$ and $x_2 = 1 / (\num{1e-6}+ .1)$.
That is, while $b$ is in fact in the range of $A$, it is not what we refer to as ``easily'' in the range of $A$.
Since the columns of $A$ are mostly orthogonal to $b$, the exact solution shown in Figure \ref{fig:motivation0} Right overdials both columns producing erroneously large canceling in the horizontal direction in order to edge up vertically even a small amount.
Ideally, one would consider these columns of $A$ unrepresentative of $b$ and instead find a control parameter that moves vertically.
Practitioners often claim that regularization handles these issues.
However, regularized least squares of the form $\text{min}_x ||b - Ax||_2^2 + \lambda ||x||_2^2$ has the solution shown in Figure \ref{fig:motivation0} Right where the non-zero aspect of the regularized solution is almost entirely in the spurious horizontal direction with only an infinitesimal component in the direction of $b$; with such regularization one can only drive the spurious horizontal component to zero with no hope of moving vertically towards the solution.

\begin{figure}[t]
\centering
\begin{subfigure}{0.32\linewidth}
\includegraphics[width=\linewidth]{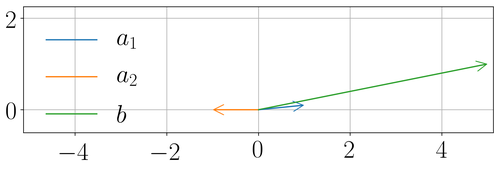}
\caption{}
\end{subfigure}
\begin{subfigure}{0.32\linewidth}
\includegraphics[width=\linewidth]{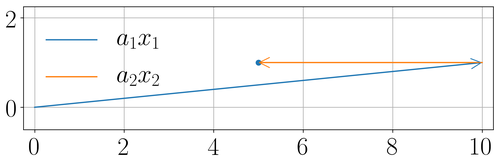}
\caption{}
\end{subfigure}
\begin{subfigure}{0.32\linewidth}
\includegraphics[width=\linewidth]{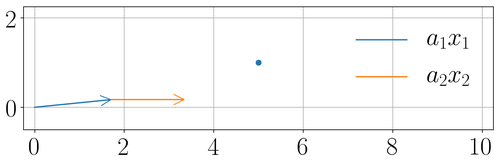}
\caption{}
\end{subfigure}
\begin{subfigure}{0.32\linewidth}
\includegraphics[width=\linewidth]{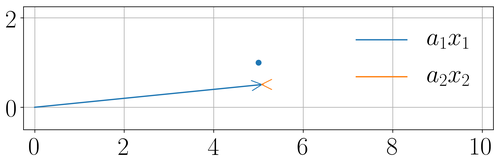}
\caption{}
\end{subfigure}
\begin{subfigure}{0.32\linewidth}
\includegraphics[width=\linewidth]{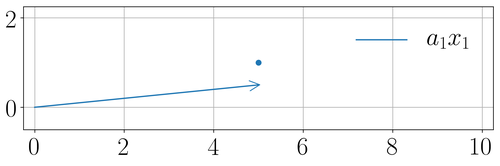}
\caption{}
\end{subfigure}
\begin{subfigure}{0.32\linewidth}
\includegraphics[width=\linewidth]{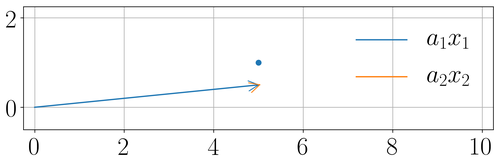}
\caption{}
\end{subfigure}
\caption{(a) A visualization of $A$'s columns and $b$ for the linear problem $Ax = b$ from Section \ref{sec:motivation} when $b = \protect\begin{bmatrix} 5 & 1\protect\end{bmatrix}^T$.
(b) The exact solution depicted by $a_1 x_1$ and $a_2 x_2$.
(c) The regularized solution with $\lambda = 1$.
(d) The regularized solution with $\lambda_1 = 0$ and $\lambda_2 = 1$.
(e) The solution when solving for $x_1$ only.
(f) The solution using coordinate descent using the MBI selection rule after a few iterations.}
\label{fig:motivation1}
\end{figure}

Next consider the case where $b = \begin{bmatrix} 5 & 1 \end{bmatrix}^T$ and the $b_1 = 5$ component is more readily captured by the columns of $A$.
The exact solution Figure \ref{fig:motivation1}b is an improvement over Figure \ref{fig:motivation0} Right since less of $a_1$ is wasted to cancel all of $a_2$, and similarly, the regularized solution shown in Figure \ref{fig:motivation1}c makes some progress towards the actual solution as compared to Figure \ref{fig:motivation0} Right.
Still, one could do better by changing the regularization in the least squares problem to have the form $\text{min}_x ||b - Ax||_2^2 + \lambda_1 x_1^2 + \lambda_2 x_2^2$ with $\lambda_1 = 0$.
Figure \ref{fig:motivation1}d shows the result for $\lambda_2 = 1$ which is highly improved.
One could do even better using only $a_1$ as shown in Figure \ref{fig:motivation1}e obtained using $\text{min}_{x_1} ||b - a_1 x_1||_2^2$.
This is equivalent to minimizing $Ax - b$ using coordinate descent after using only one column.

\section{Pruning} \label{sec:pruning}

We illustrate our approach by solving a generic nonlinear least squares optimization problem of the form $\text{min}_x ||f(x)||_2^2$ using a Gauss-Newton based method, computing the Jacobian $J = \partial f / \partial x$ to utilize the first-order Taylor expansion $f(x + \delta x) = f(x) + J \delta x$ at each iteration.
The minimization problem at the current iteration is $\text{min}_{\delta x} || f(x) + J \delta x ||_2^2$.
This is a linear least squares problem and can be solved using $J^TJ \delta x = -J^T f(x)$ to find the $\delta x$ that makes progress towards the solution.
The least squares solution can also be computed by solving $J \delta x = -f(x)$ using QR factorization; however, it is not clear that the least squares solution is necessary or desirable.

\begin{figure}[b]
\centering
\includegraphics[width=0.3\linewidth]{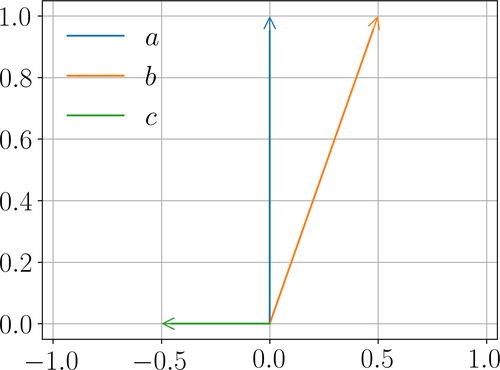}
\hfill
\includegraphics[width=0.3\linewidth]{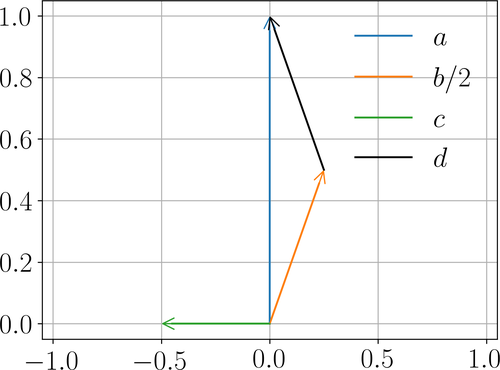}
\hfill
\raisebox{-0.25cm}{\includegraphics[width=0.24\linewidth]{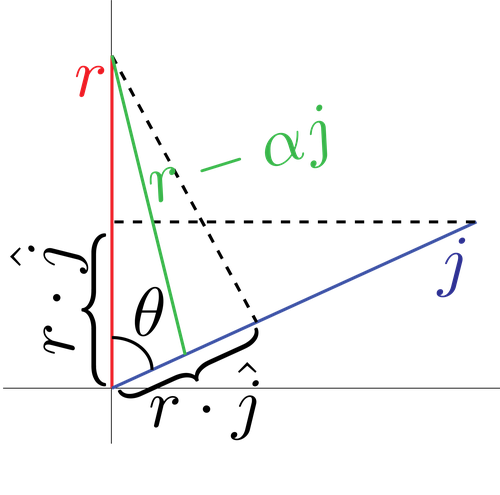}}
\caption{Left: Here, $a = b + c$, but $c$ is only valid for making progress towards $a$ in conjunction with $b$.
While using $c$ may be desirable when trying to solve a linear system of equations, using $c$ (which is perpendicular to $a$) to progress when solving a linearized high-dimensional nonlinear problem may not produce any meaningful progress.
Middle: In fact, it may be desirable to only progress in the direction of $b$ for a while, stopping before the remaining residual starts to become uncorrelated.
At that point, it would be better to find a new direction (in this case $d$) that leads back towards $a$.
Right: MBI simply minimizes $\theta$; GS and our approach instead consider $\hat{r} \cdot j$.
Unlike GS, our approach also prefers that the remaining residual $r - \alpha j$ still be correlated with $j$.
This penalizes searching too far along $j$, especially since that can lead to using uncorrelated directions as corrections (see the left subfigure).
Instead, we stop progress along $j$ while the remaining residual is still correlated, enabling us to seek more correlated directions to continue progressing, similar to LARS \cite{efron2004least}.
}
\label{fig:orthogonal_correction}
\end{figure}

We thus propose an alternative approach to solving $J \delta x = -f(x)$ which depends on computing the correlation between each column $j_i$ of $J$ and $-f(x)$.
Similarly to LARS \cite{efron2004least} and MBI \cite{chen2012maximum}, we use the dot product magnitude $|\hat{j}_i \cdot f(x)|$ to determine correlation, where $\hat{j}_i = j_i / \| j_i \|_2$ .
We first prune out columns $j_i$ that are poorly correlated with $f$ as these $j_i$ make no progress towards the solution in the linearized state of the problem.
Furthermore, these columns can only act as corrections on actual progress in conjunction with other columns; however, these corrections are only valid when the problem is truly linear.
In the case of a nonlinear problem, it is unclear whether these uncorrelated corrections help to minimize the nonlinear $||f(x)||_2^2$.
See Figure \ref{fig:orthogonal_correction} Left/Middle.
Motivated by the GS rule, one might instead prune using the dot product via $|j_i \cdot f(x)|$ to consider large decreases in residual with small steps; however, this may leave columns that are large and poorly correlated unpruned.
Thus, we use the absolute dot product of the normalized vectors as the correlation measure for its geometric interpretation as the absolute cosine of the angle between $j_i$ and $f(x)$.

Pruning columns of $J$ to get a reduced $J_S$ has the additional benefit of potentially eliminating portions of the null space of $J$, as the pruned out columns may not have full rank or a combination of the pruned out columns with the remaining columns may not have full rank; this pruning may also improve the condition number.
This is especially prudent when working with a large number of dimensions in which case the dimension of the null space of $J$ and the condition number of $J$ may be large.
Moreover, this problem is exacerbated when regularization is not used.
At this point, one could compute a modified step $\delta x_S$ that uses the columns of $J_S$ to make progress towards $-f(x)$.
However, if $J_S$ contains a null space, some regularization would still be needed to obtain a reasonable solution.

\section{Solving the Pruned System} \label{sec:solving}

To avoid regularization, we pursue a coordinate descent strategy to solve $J_S \delta x_S = -f(x)$.
At each iteration of the algorithm, a single column $j_i$ of the Jacobian is used to make progress towards $-f(x)$.
Since only one column is examined at any given iteration, we do not have to use regularization to avoid null spaces.
There are a few strategies for choosing the column to update that follow our motivation of using correlated columns.
For example, one could choose the column with the largest correlation with the residual or choose the column which results in the largest decrease in the residual similar to the MBI rule.
Motivated by the GS rule \cite{nutini2015coordinate}, we instead choose the column that maximizes the reduction in residual for a given step, which in effect balances correlation with residual reduction.
Assuming the residual is measured using the squared L2 norm, we choose the column $j_i$ maximizing
\begin{align}
\frac{\Delta (r^T r)}{\Delta \alpha_i} & = \frac{||r(\delta x_S)||_2^2 - ||r(\delta x_S) - \alpha_i j_i ||_2^2}{|\alpha_i|}
\label{eq:our_gauss_southwell}
\end{align}
where $r(\delta x_S) = -f(x) - J_S \delta x_S$ is the residual for the linearized problem and $\alpha_i$ is the step size when $j_i$ is the coordinate direction.
Flipping all $j_i$ so that $r(\delta x_S)^T j_i > 0$ leads to $\alpha_i > 0$, which in turn means that finding the column $j_i$ that maximizes Equation \ref{eq:our_gauss_southwell} is equivalent to the column $j_i$ that maximizes $M = 2 r(\delta x_S)^T j_i - \alpha_i ||j_i||_2^2$.

When $\alpha_i$ is chosen to make maximal progress, i.e.\ $\alpha_i j_i = (r(\delta x_S) \cdot \hat{j}_i) \hat{j}_i$ or $\alpha_i = r(\delta x_S)^T j_i / \| j_i \|_2^2$, we obtain $M = r(\delta x_S)^T j_i$ which is standard for GS.
Similarly, as $\alpha_i \to 0$, $M \to  2 r(\delta x_S)^T j_i$.
Geometrically, this maximizes the projection of $j_i$ into the direction of $r(\delta x_S)$, i.e.\ $j_i \cdot \hat{r}(\delta x_S)$, since $\| r(\delta x_S) \|$ does not vary when considering which $j_i$ to use.
In comparison, the MBI rule chooses the $j_i$ that maximizes $\hat{j}_i \cdot \hat{r}(\delta x_S)$ ignoring the magnitude of $j_i$, which unfortunately allows for directionally correlated but small magnitude columns that may require large overdialed $\alpha_i$ to make progress.

When more conservatively choosing $\alpha_i$ less than the greedy value to make the maximal progress, $\alpha_i < r(\delta x_S)^T j_i / \| j_i \|_2^2$ and $M = r(\delta x_S)^T j_i + r(\delta x_S)^T j_i - \alpha_i ||j_i||_2^2$.
Whereas the greedy choice eliminated the last two terms in $M$ giving a result in line with GS, our approach augments the GS-like term with $r(\delta x_S)^T j_i - \alpha_i ||j_i||_2^2 = (r(\delta x_S) - \alpha_i j_i) \cdot j_i$ which compares the remaining residual to the search direction $j_i$.
Maximizing this latter term helps to prefer search directions that remain correlated with the new residual preventing uncorrelated gains that might later be removed using uncorrelated directions as in Figure \ref{fig:orthogonal_correction} Left.
See Figure \ref{fig:orthogonal_correction} Right.
We note that one may replace our proposed Heaviside pruning (in Section \ref{sec:pruning}) with a differentiable soft-pruning penalty in Equation \ref{eq:our_gauss_southwell}.

We generally only execute a few iterations of coordinate descent using these update rules to mimic the regularization effects of early stopping \cite{goodfellow2016deep} and truncated-Newton methods \cite{nash2000survey} as it prevents overfitting to the current linear model.
Furthermore, we also terminate the coordinate descent iterations early if the decrease in L2 error is low.
Stopping early also prevents us from reaching the undesirable least squares solution as in LARS \cite{efron2004least}.

\section{Experiments} \label{sec:experiments}

Consider matching hand-drawn rotoscope curves on a captured image to the corresponding two-dimensional projections of similar curves drawn on a three-dimensional face model.
The face surface $x(w)$ is driven by blendshapes \cite{lewis2014practice} and linear blend skinning for the jaw; let $w$ represent the full set of controls for the jaw angles, jaw translation, and face blendshapes.
Assuming the pointwise correspondences between the curves are known, we solve a nonlinear least squares problem
\begin{align}
    \text{min}_w \| C^* - C(x(w)) \|_2^2 \label{eq:roto}
\end{align}
to estimate $w$, where $C^*$ are the points on the rotoscope curves and $C(x(w))$ are the corresponding points from the face model projected into the image plane.
Here, we assume that the camera parameters and the rigid alignment are precomputed.

To solve this problem, practitioners often add L2 regularization of the parameters (e.g.\ $\lambda \| w\|_2^2$).
Although this achieves reasonable results when looking at the geometry and/or synthetic render, the resulting parameter values $w$ are typically densely activated, making it nearly impossible to utilize those values for editing or for extracting semantic information.

\subsection{Synthetic Tests} \label{sec:synth_tests}

\begin{figure}[hbp]
\centering
\begin{subfigure}[t]{\dimexpr0.14\linewidth+20pt\relax}
    \makebox[20pt]{\raisebox{35pt}{\rotatebox[origin=c]{90}{\small Dogleg}}}%
    \includegraphics[width=\dimexpr\linewidth-20pt\relax]{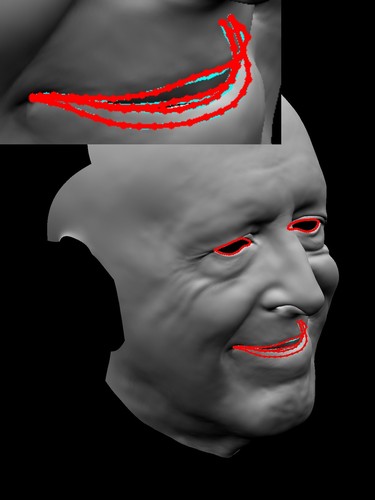}
    \makebox[20pt]{\raisebox{35pt}{\rotatebox[origin=c]{90}{\small Dogleg + L2}}}%
    \includegraphics[width=\dimexpr\linewidth-20pt\relax]{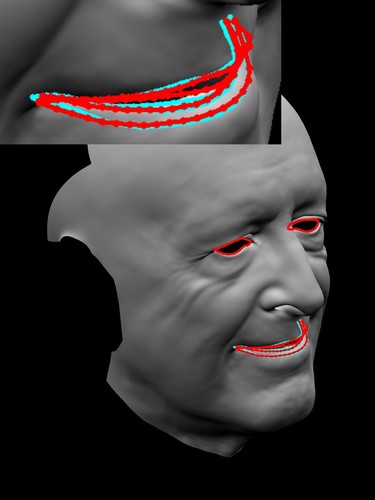}
    \makebox[20pt]{\raisebox{35pt}{\rotatebox[origin=c]{90}{\small BFGS + Soft L1}}}%
    \includegraphics[width=\dimexpr\linewidth-20pt\relax]{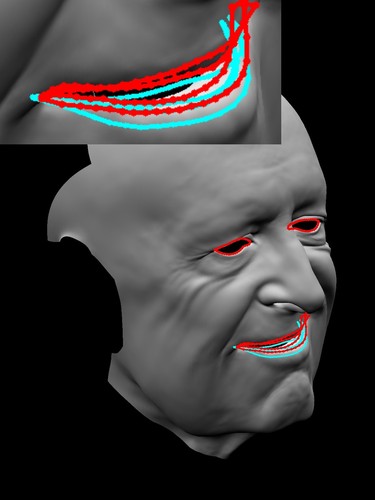}
    \makebox[20pt]{\raisebox{35pt}{\rotatebox[origin=c]{90}{\small Our Approach}}}%
    \includegraphics[width=\dimexpr\linewidth-20pt\relax]{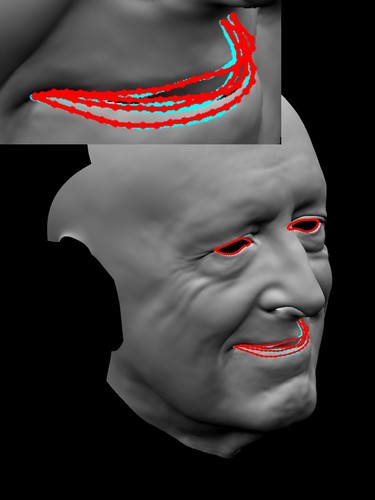}
    \caption{No Noise}
    \label{fig:synthetic_mesh_comparison_a}
\end{subfigure}
\begin{subfigure}[t]{0.14\linewidth}
    \includegraphics[width=\linewidth]{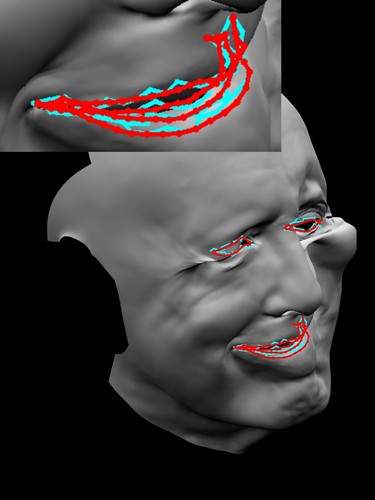}
    \includegraphics[width=\linewidth]{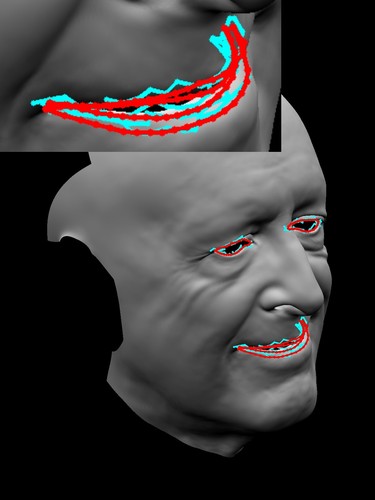}
    \includegraphics[width=\linewidth]{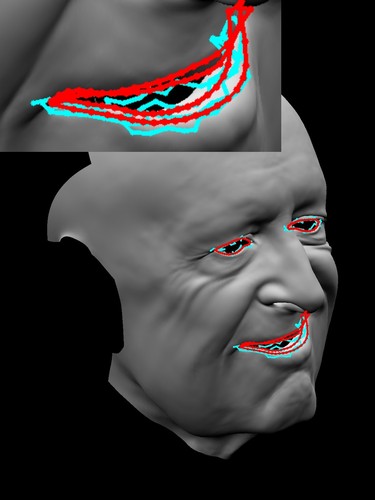}
    \includegraphics[width=\linewidth]{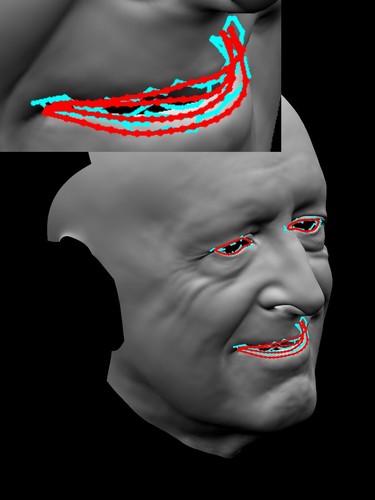}
    \caption{(\num{0.005})}
\end{subfigure}
\begin{subfigure}[t]{0.14\linewidth}
    \includegraphics[width=\linewidth]{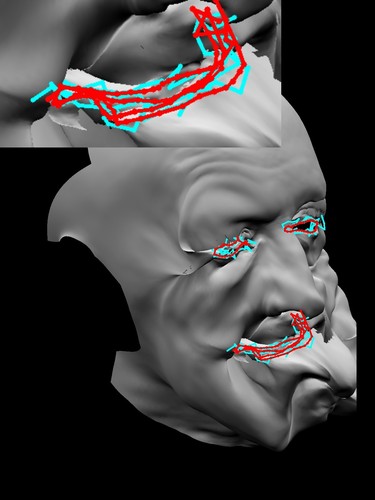}
    \includegraphics[width=\linewidth]{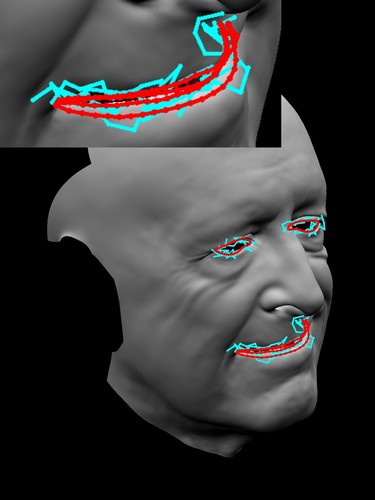}
    \includegraphics[width=\linewidth]{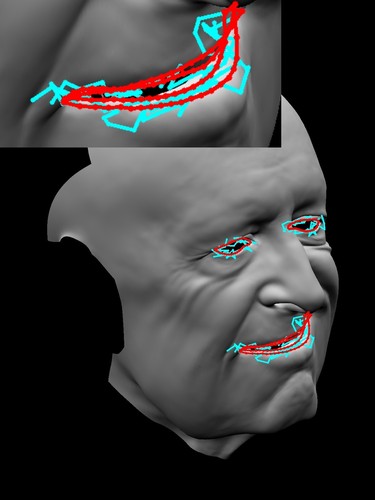}
    \includegraphics[width=\linewidth]{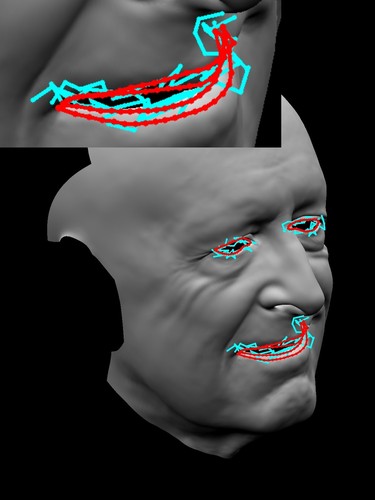}
    \caption{(\num{0.01})}
\end{subfigure}
\begin{subfigure}[t]{0.16\linewidth}
    \includegraphics[width=\linewidth]{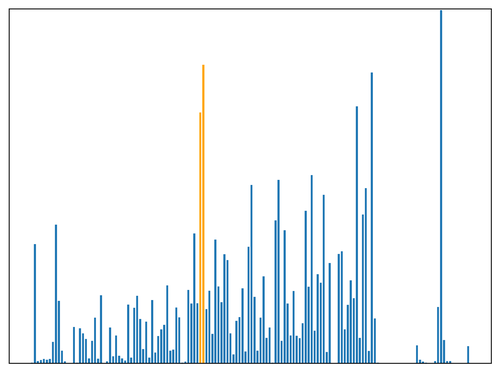}\\
    \vspace*{17pt}\\
    \includegraphics[width=\linewidth]{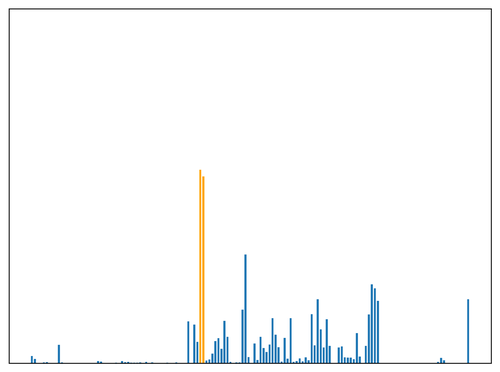}\\
    \vspace*{16pt}\\
    \includegraphics[width=\linewidth]{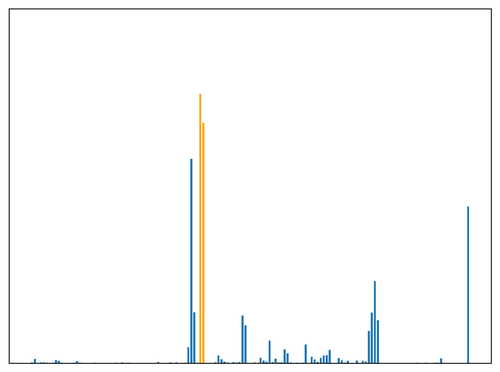}\\
    \vspace*{16pt}\\
    \includegraphics[width=\linewidth]{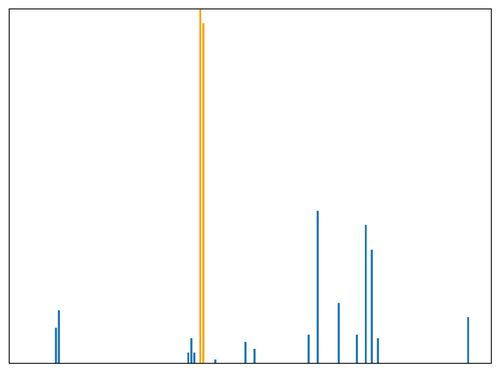}\\
    \vspace*{-13pt}
    \caption{No Noise}
\end{subfigure}
\begin{subfigure}[t]{0.16\linewidth}
    \includegraphics[width=\linewidth]{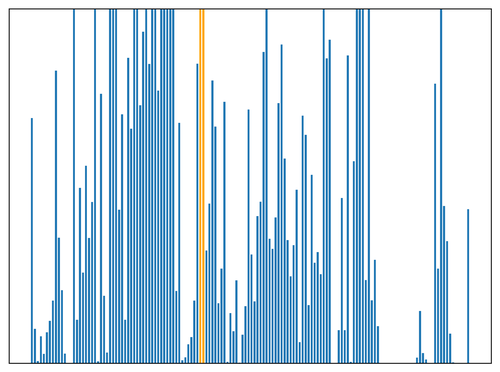}\\
    \vspace*{17pt}\\
    \includegraphics[width=\linewidth]{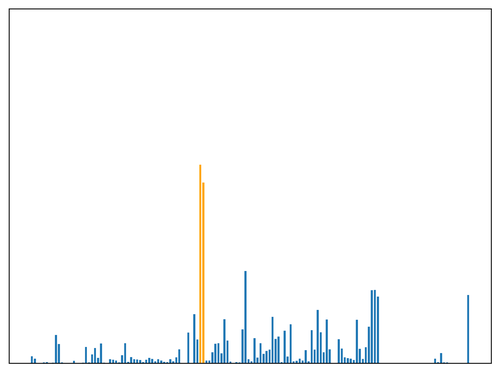}\\
    \vspace*{16pt}\\
    \includegraphics[width=\linewidth]{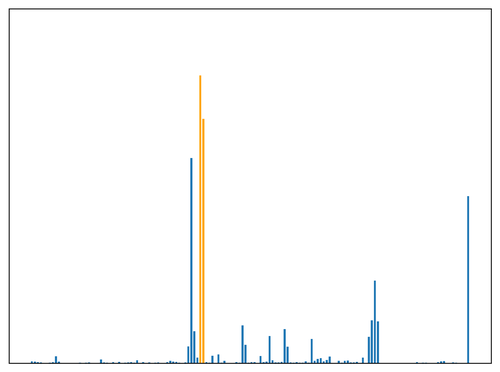}\\
    \vspace*{16pt}\\
    \includegraphics[width=\linewidth]{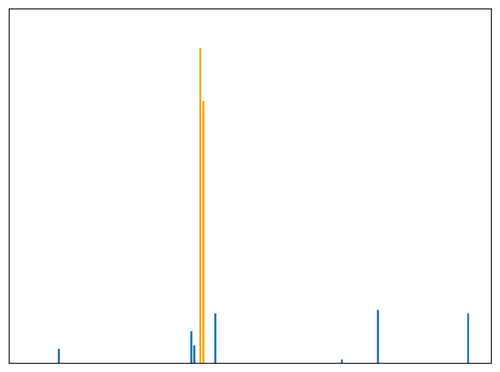}\\
    \vspace*{-13pt}
    \caption{Noise (\num{0.005})}
\end{subfigure}
\begin{subfigure}[t]{0.16\linewidth}
    \includegraphics[width=\linewidth]{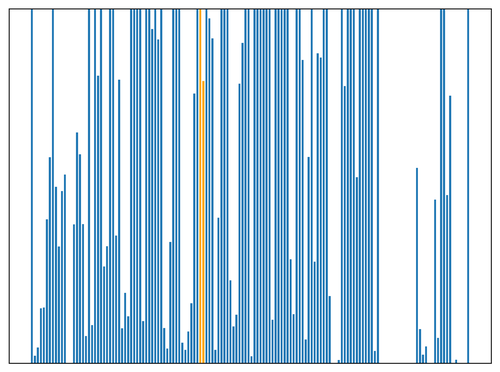}\\
    \vspace*{17pt}\\
    \includegraphics[width=\linewidth]{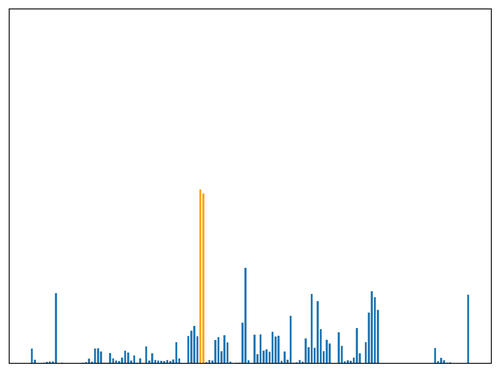}\\
    \vspace*{16pt}\\
    \includegraphics[width=\linewidth]{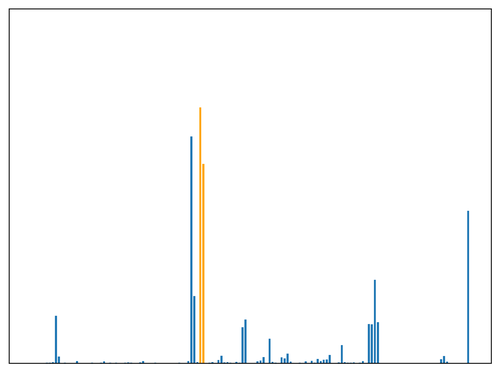}\\
    \vspace*{16pt}\\
    \includegraphics[width=\linewidth]{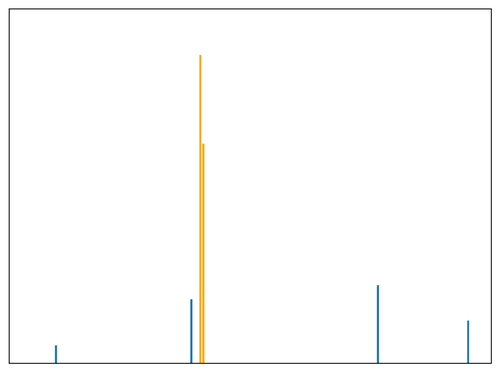}\\
    \vspace*{-13pt}
    \caption{Noise (\num{0.01})}
\end{subfigure}
\caption{Left: As we increase the amount of noise added to the points on the blue target curve, the Dogleg method without regularization overfits causing the mesh to ``explode'' in spite of having the smallest error as measured by Equation \ref{eq:roto} (typical of overfitting).
On the other hand, the standard regularization terms and our approach prevent the model from overfitting to the noisy curves.
Right: The corresponding blendshape weights.
The target solution was generated by setting the two orange columns to one and the blue columns to zero.
The figure heights are clipped at \num{1.0} and many parameter values exceed that.
Though the reguarized solves have smaller, spurious weights than the non-regularized version (second and third row vs.~first row), our approach (last row) produces a much sparser solution with more semantic meaning even in the presence of noise.}
\label{fig:synthetic_mesh_comparison}
\end{figure}

We create a $w^*$ with two non-zero components (value \num{1.0}) corresponding to the smile expression to produce a face surface $x(w^*)$ from which the barycentrically embedded contour points can be projected into the image plane, generating a synthetic target $C^*$.
We then solve Equation \ref{eq:roto} using Dogleg with no prior, Dogleg with a prior weight of $\lambda = 3600$, and BFGS \cite{nocedal2006numerical} with a soft-L1 prior with a weight of $3600$ (i.e.\ with an extra term $3600 \sum_i 2 (\sqrt{1 + w_i^2} -1)$ \cite{charbonnier1997deterministic}).
For our approach, we firstly prune all columns of the Jacobian whose angle to the residual has an absolute cosine less than $\num{0.3}$ (determined experimentally).
Then, $\alpha_i$ is set to a fixed size with $\tau = \num{1e-2}$ and coordinate descent is run until the linear L2 error no longer sufficiently decreases or when over $10$ coordinates get used.
We limit all four methods to $\leq 10$ Gauss-Newton iterations; Figure \ref{fig:synthetic_mesh_comparison_a} shows the results.

Next, we add an increasing amount of uniformly distributed noise to the image locations of the targeted contour points.
As expected, Dogleg with no regularization produces reasonable results when the noise is low but begins to overfit to the errorneous data as the amount of noise increases.
See Figure \ref{fig:synthetic_mesh_comparison}.
Both of the regularized approaches as well as our approach, however, are able to target the noisy curves (without overfitting) producing more reasonable geometry.
The right portion of Figure \ref{fig:synthetic_mesh_comparison} shows our approach yields sparser (more semantic) solutions even with noise.
Tables \ref{tbl:synth1} and \ref{tbl:synth2} demonstrate quantitative success of our method for these examples (in both error and sparsity).

\begin{table}[!ht]
\caption{Comparing accuracy of estimating the original blendshape weights in the synthetic tests under various metrics. Our method produces the lowest error regardless of noise and metric.}
\setlength\tabcolsep{2.5pt}
\begin{tabularx}{\linewidth}{Xrrrcrrrcrrr}
    \toprule
    & \multicolumn{3}{c}{L2 Error} & \phantom{a} & \multicolumn{3}{c}{L1 Error} & \phantom{a} & \multicolumn{3}{c}{EMD \cite{710701} Error} \\
    \cmidrule{2-4} \cmidrule{6-8} \cmidrule{10-12}
    Method & No Noise & \num{0.005} & \num{0.01} & & No Noise & \num{0.005} & \num{0.01} & & No Noise & \num{0.005} & \num{0.01} \\
    \midrule
    Dogleg & $2.578$ & $8.815$ & $20.325$ & & $19.227$ & $70.852$ & $157.22$ & & $0.128$ & $0.485$ & $1.07$ \\
    Dogleg+L2 & $0.972$ & $0.952$ & $1.209$ & & $4.954$ & $5.324$ & $5.704$ & & $0.034$ & $0.036$ & $0.039$ \\
    BFGS+Soft L1 & $0.923$ & $0.91$ & $1.023$ & & $3.139$ & $3.057$ & $3.359$ & & $0.0215$ & $0.021$ & $0.023$ \\
    Our & $\textbf{0.741}$ & $\textbf{0.392}$ & $\textbf{0.509}$ & & $\textbf{2.208}$ & $\textbf{0.99}$ & $\textbf{1.08}$ & & $\textbf{0.015}$ & $\textbf{0.007}$ & $\textbf{0.007}$  \\
    \bottomrule
    \end{tabularx}
\label{tbl:synth1}
\end{table}

\begin{table}[!ht]
\caption{The sparsty of results on the synthetic tests using common sparsity metrics (larger number is better) \cite{hurley2009comparing}.
The $l^0$ metric counts how many weights are strictly $0$.
The Gini metric is $1 - 2 \sum_{k=1}^ \frac{c_k}{||c||_1} (\frac{N - k + 0.5}{N})$ where $c$ is the sorted weights and $c_k$ is the $k$\textsuperscript{th} largest blendshape weight.}
\begin{tabularx}{\linewidth}{Xrrrcrrr}
    \toprule
    & \multicolumn{3}{c}{$l^0$ Metric} & \phantom{a} & \multicolumn{3}{c}{Gini Metric} \\
    \cmidrule{2-4} \cmidrule{6-8}
    Method & No Noise & \num{0.005} & \num{0.01} & & No Noise & \num{0.005} & \num{0.01}  \\
    \midrule
    Dogleg & $21$ & $21$ & $21$ & & $0.628$ & $0.580$ & $0.607$ \\
    Dogleg+L2 & $21$ & $21$ & $21$ & & $0.807$ & $0.745$ & $0.739$  \\
    BFGS+Soft L1 & $21$ & $21$ & $21$ & & $0.913$ & $0.905$ & $0.916$ \\
    Our & $\textbf{128}$ & $\textbf{137}$ & $\textbf{140}$ & & $\textbf{0.949}$ & $\textbf{0.974}$ & $\textbf{0.978}$ \\
    \bottomrule
\label{tbl:synth2}
\end{tabularx}
\end{table}

\subsection{Real Image Data} \label{sec:real_tests}

For each captured image, an artist draws eight curves around the mouth and eyes.
Another artist hand-selects the corresponding contours on the three-dimensional face model which are projected into the image plane using calibrated camera extrinsic and intrinsic parameters \cite{heikkila1997four}.
Generally, these two sets of contours will contain a different number of control points; as a result, we uniformly resample both in the image plane.
The uniformly sampled curves are then used in Equation \ref{eq:roto}.

\begin{figure*}[htpb]
\centering
\begin{subfigure}[b]{0.17\linewidth}
    \includegraphics[width=\linewidth]{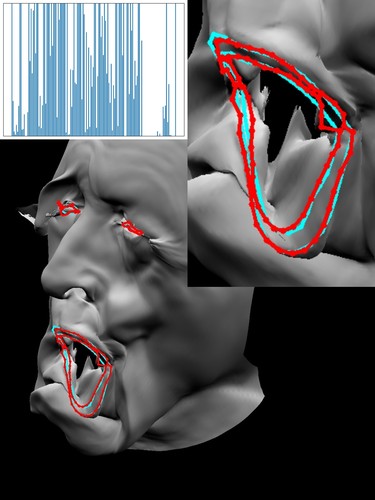}
    \caption{Dogleg}
\end{subfigure}
\begin{subfigure}[b]{0.17\linewidth}
    \includegraphics[width=\linewidth]{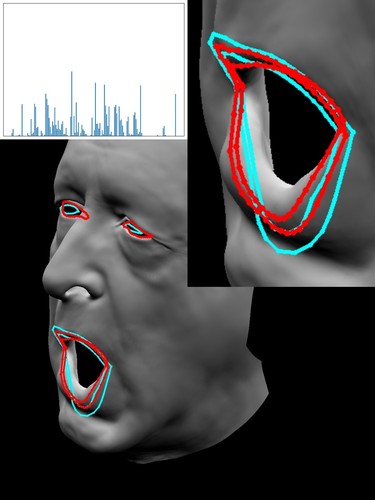}
    \caption{Dogleg+L2}
\end{subfigure}
\begin{subfigure}[b]{0.17\linewidth}
    \includegraphics[width=\linewidth]{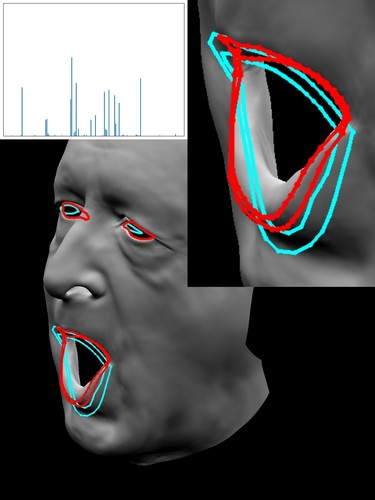}
    \caption{BFGS+Soft L1}
\end{subfigure}
\begin{subfigure}[b]{0.17\linewidth}
    \includegraphics[width=\linewidth]{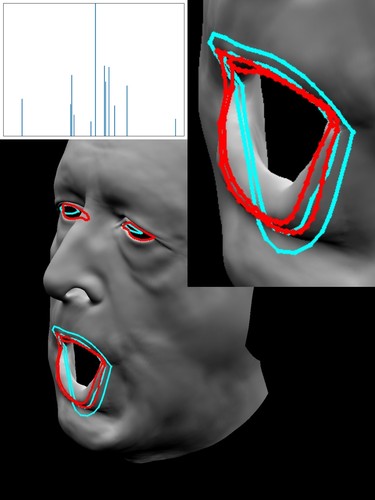}
    \caption{Our Approach}
\end{subfigure}
\begin{subfigure}[b]{0.17\linewidth}
    \includegraphics[width=\linewidth]{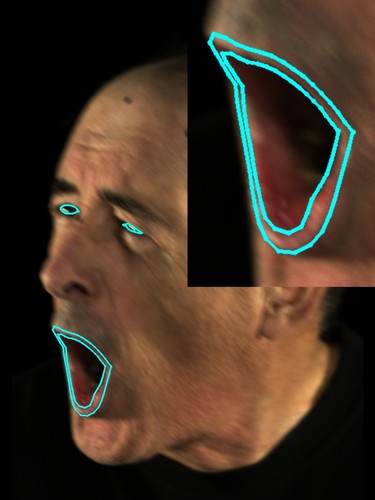}
    \caption{Target}
\end{subfigure}
\hfill
\caption{
Dogleg without regularization clearly overfits to the curves, producing highly unrealistic face shapes.
Dogleg with regularization performs better but sometimes overfits as well.
This could be tuned by increasing the regularization weight at the cost of potentially damping out the performance.
Our approach produces facial expressions that are reasonably representative of the captured image.
The inset bar plots demonstrate the sparsity of the weights for each of the methods.
Our method generally produces the sparsest set of weights; e.g.\ in this frame (\num{1142}), our method has \num{12} non-zero parameter values while L2 regularization produces fully dense results and soft L1 regularization has \num{49} significant parameter values (i.e.\ $> \num{1e-3}$).  See appendix for additional frames.}
\label{fig:plate_mesh_comparison}
\end{figure*}

Our approach and using regularization give the most reasonable geometric results---similar to the synthetic case.
However, in certain frames, our approach seems to over-regularize the result and cause the mouth to not quite hit the desired expression.
This could be due, in part, to not having a fully accurate rigid alignment.
However, a major benefit of our method is the resulting sparsity in the weights.
While the Dogleg method with and without regularization will dial in nearly all the parameters to a non-zero value, our approach generally dials in a small number of weights.
See Figure \ref{fig:plate_mesh_comparison}.
Similarly, the soft L1 regularized solution also produces sparser solutions than the L2 regularized solution; however, due to approximations in the chosen optimization approach (BFGS, Soft L1), it produces many small (i.e.\ $< \num{1e-3}$) weights instead of zero values.
While one could use a threshold to clamp values to zero, care must be taken to not accidentally threshold small blendshape weights that contribute significantly to the overall performance.
We note that one could potentially use iteratively reweighted least squares methods \cite{chartrand2008iterative} to obtain sparse solutions.

\subsection{Parameter Study}

\textbf{Step Size:}
Recall $\alpha_i$ are the parameters, i.e.\ blendshape weights, that generally vary between \num{0} and \num{1}; thus we choose to compare fixed step sizes of $\tau = 0.01, 0.02, 0.1, 0.5$ and \num{1.0} to the full, greedy step, i.e.\ $\alpha_i = r^T j_i / \| j_i \|_2^2$.
Without pruning, we run \num{10} Gauss-Newton iterations with no thresholding for the relative decrease in L2 error and an upper limit of \num{10} unique coordinates used.
We find that smaller step sizes achieve better overall facial shapes (see appendix) and less overdialed weights.
In particular, the greedy step dials \num{7} weights to be greater than $1$ while the $\tau = 0.02$ and $\tau = 0.01$ step sizes only dial in \num{4} such weights.
Removing the eye rotoscope curves causes the overdialed weights to disappear; however, as seen in the appendix, the greedy step causes the mouth to move unnaturally.

We also compare the effect of using fixed step sizes in Equation \ref{eq:our_gauss_southwell} versus the full, greedy step size equivalent to GS without pruning.
To isolate this variable, we run \num{10} Gauss-Newton iterations with no thresholding for the relative decrease in L2 error and an upper limit of \num{10} unique coordinates used.
We vary $\alpha_i$ in Equation \ref{eq:our_gauss_southwell} but set the actual step size taken to be fixed at \num{0.01}.
As shown in the appendix, while the resulting geometry and weights are all similar, our approach of allowing the step size to influence the chosen coordinate allows the optimization to more quickly reduce the error in earlier Gauss-Newton iterations.

\textbf{Pruning:}
\begin{figure}[t]
\centering
\begin{subfigure}[b]{0.22\linewidth}
    \includegraphics[width=\linewidth]{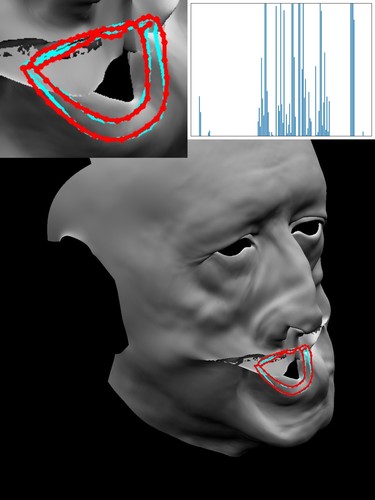}
    \caption{No Pruning}
\end{subfigure}
\begin{subfigure}[b]{0.22\linewidth}
    \includegraphics[width=\linewidth]{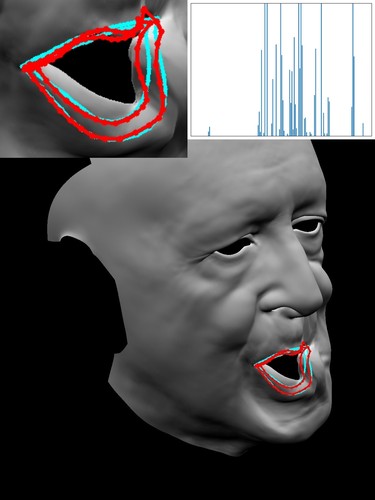}
    \caption{\num{0.2}}
\end{subfigure}
\begin{subfigure}[b]{0.22\linewidth}
    \includegraphics[width=\linewidth]{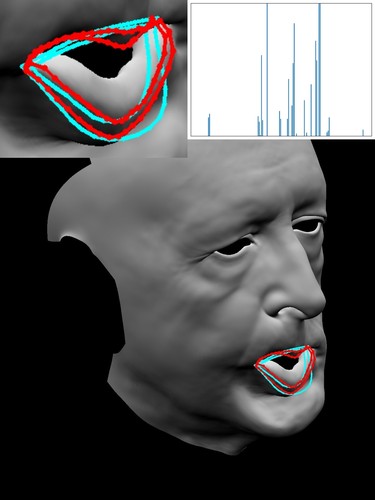}
    \caption{\num{0.3}}
\end{subfigure}
\begin{subfigure}[b]{0.22\linewidth}
    \includegraphics[width=\linewidth]{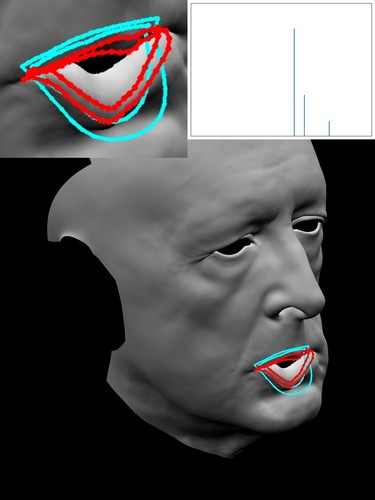}
    \caption{\num{0.5}}
\end{subfigure}
\caption{As we increase the minimum absolute dot product threshold for pruning, the resulting solution becomes sparser and more regularized.}
\label{fig:pruning_parameter_comparison}
\vspace{-2.5mm}
\end{figure}
We use the correlation metric of $\hat{r}^T \hat{j}_i$ from LARS and MBI to prune poorly correlated coordinate directions.
We choose threshold values of \num{0.0} (no pruning), \num{0.2}, \num{0.3}, and \num{0.5} and run \num{10} Gauss-Newton iterations with a step size of \num{0.01} with no thresholding for the relative decrease in L2 error.
To emphasize the effect of pruning, we allow up to \num{50} unique coordinates per linearization, and focus only on the rotoscope curves around the mouth.
With little to no pruning the model overfits and the geometry around the mouth deforms unreasonably.
As the pruning threshold increases, the geometry becomes more regularized and the blendshape weights are less overdialed and sparser as the optimization is forced to use only the most correlated directions.
See Figure \ref{fig:pruning_parameter_comparison}.
However, we caution that too much pruning causes MBI style column choices.

\section{Conclusion}

Taking the linearization of the nonlinear problem as a mere suggestion for a search strategy, we prune uncorrelated directions, pursue a coordinate descent approach that does not need regularization, and choose search directions in order to maximize gains in reducing the residual with minimal parameter value increases.
We stress that this is more generally an outline of an improved search strategy as opposed to a formal and precise new method; yet, our first attempts at such an approach led to highly improved results.
In the case of estimating three-dimensional facial expression from a mere eight contours drawn on a single two-dimensional RGB image, we were able to robustly estimate clean, sparse parameter values with good semantic meaning in a highly underconstrained situation where one would typically need significant regularization.
In fact, the standard approach without regularization was widly inaccurate, and while regularization helped to moderate the overall face shape, it excited almost every parameter in the model clouding semantic interpretation.

As future work, we would be interested in considering similar search strategies for training neural networks with limited data.
Such situations are similar to our example where we only use sparse rotoscope curves that are only suggestive of the desired expression, instead of dense data for every triangle (such as when using shape-from-shading or optical flow).

\newpage

\section*{Broader Impact}

The present work is focused on nonlinear optimization, which means there is potential impact for the improved design and training of neural networks and the associated consequences thereof.  In the specific use case we considered, using our search strategy to infer sparser and more interpretable parameters for a facial model from image data, one could imagine impact in disparate fields such as computer animation (for feature films, video games, etc.) and surveillance.

\begin{ack}
Research supported in part by ONR N00014-13-1-0346, ONR N00014-17-1-2174, ARL AHPCRC W911NF-07-0027, and generous gifts from Amazon and Toyota.
In addition, we would like to thank both Reza and Behzad at ONR for supporting our efforts into computer vision and machine learning, as well as Cary Phillips and Industrial Light \& Magic for supporting our efforts into facial performance capture.
M.B. was supported in part by The VMWare Fellowship in Honor of Ole Agesen.
We would also like to thank Paul Huston for his acting.
\end{ack}

{\small
\bibliographystyle{ieee}
\bibliography{biblio}
}

\appendix

\section{Additional Real Image Frames}

\begin{figure*}[htpb]
\centering
\begin{subfigure}[b]{\dimexpr0.17\linewidth+10pt\relax}
    \makebox[10pt]{\raisebox{60pt}{\rotatebox[origin=c]{90}{1124}}}%
    \includegraphics[width=\dimexpr\linewidth-10pt\relax]{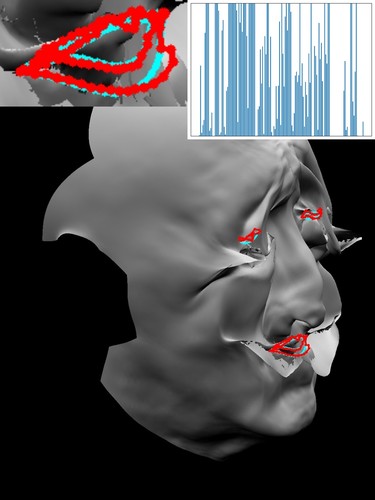}
    \makebox[10pt]{\raisebox{60pt}{\rotatebox[origin=c]{90}{1151}}}%
    \includegraphics[width=\dimexpr\linewidth-10pt\relax]{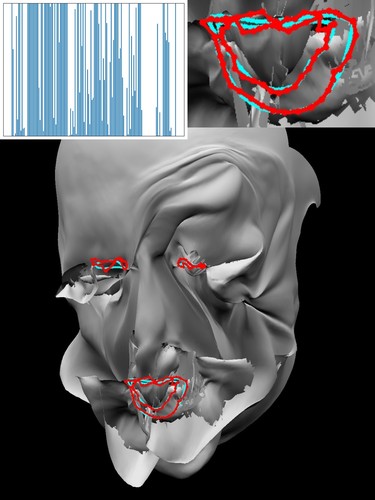}
    \makebox[10pt]{\raisebox{60pt}{\rotatebox[origin=c]{90}{1167}}}%
    \includegraphics[width=\dimexpr\linewidth-10pt\relax]{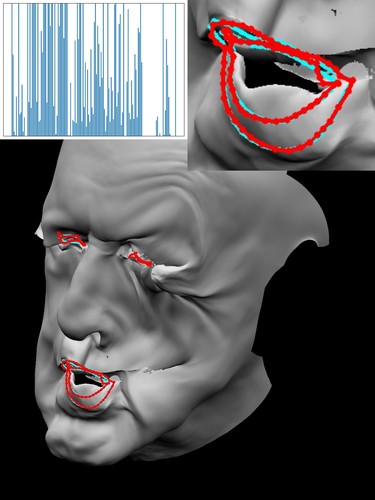}
    \makebox[10pt]{\raisebox{60pt}{\rotatebox[origin=c]{90}{1175}}}%
    \includegraphics[width=\dimexpr\linewidth-10pt\relax]{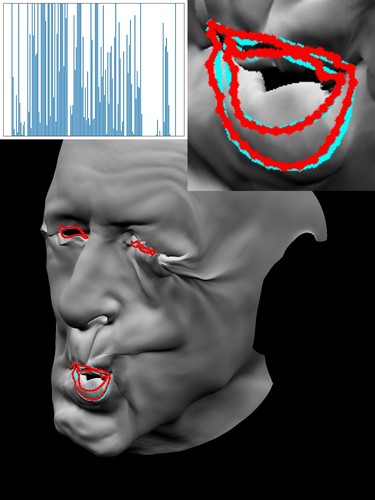}
    \caption{Dogleg}
\end{subfigure}
\begin{subfigure}[b]{0.17\linewidth}
    \includegraphics[width=\linewidth]{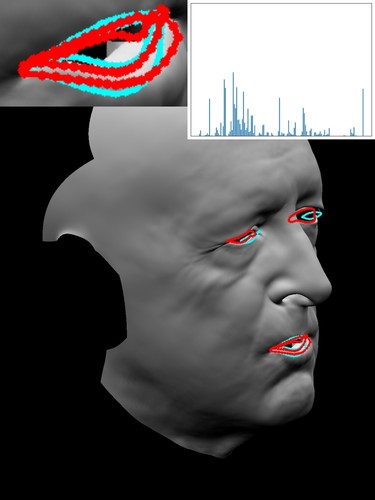}
    \includegraphics[width=\linewidth]{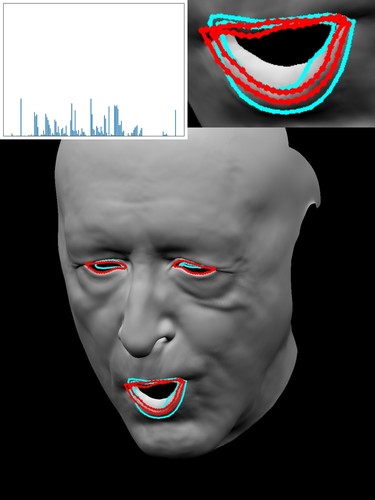}
    \includegraphics[width=\linewidth]{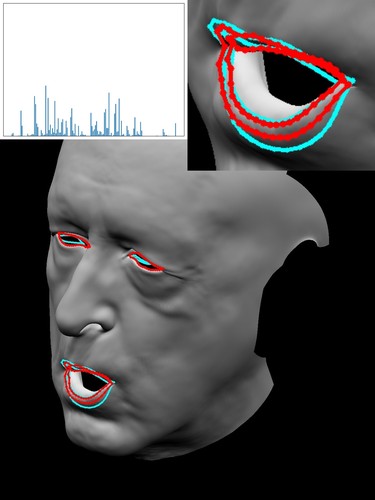}
    \includegraphics[width=\linewidth]{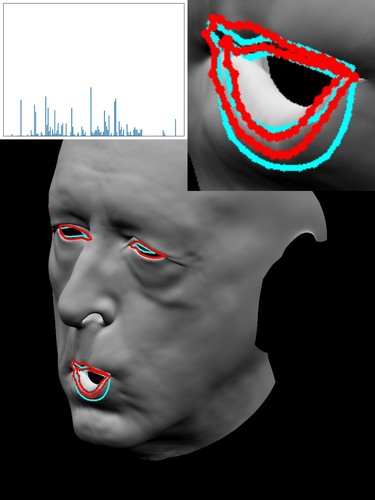}
    \caption{Dogleg+L2}
\end{subfigure}
\begin{subfigure}[b]{0.17\linewidth}
    \includegraphics[width=\linewidth]{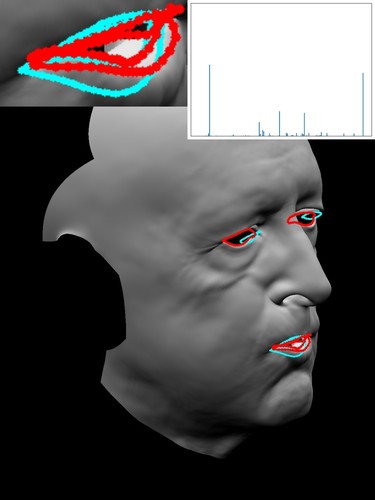}
    \includegraphics[width=\linewidth]{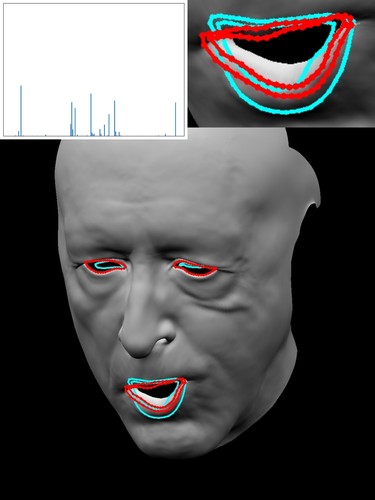}
    \includegraphics[width=\linewidth]{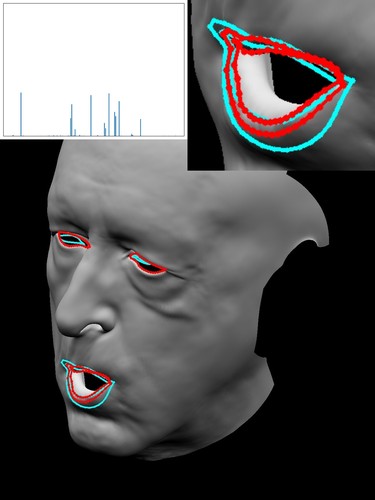}
    \includegraphics[width=\linewidth]{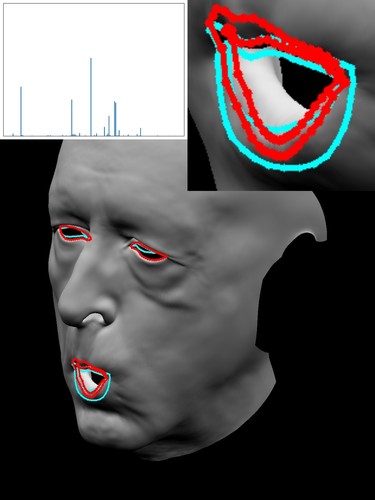}
    \caption{BFGS+Soft L1}
\end{subfigure}
\begin{subfigure}[b]{0.17\linewidth}
    \includegraphics[width=\linewidth]{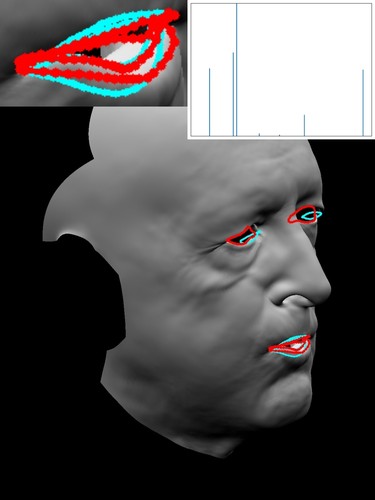}
    \includegraphics[width=\linewidth]{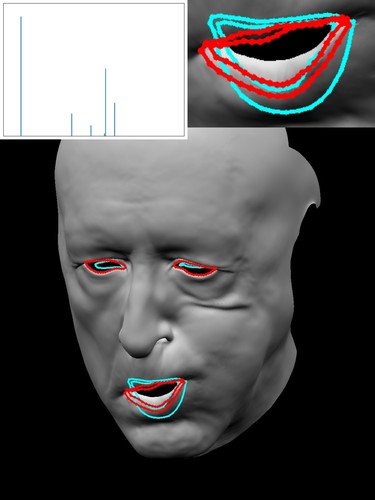}
    \includegraphics[width=\linewidth]{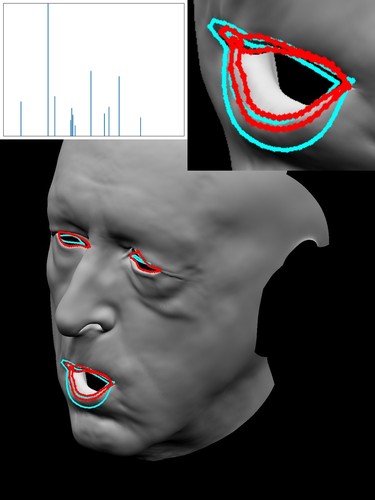}
    \includegraphics[width=\linewidth]{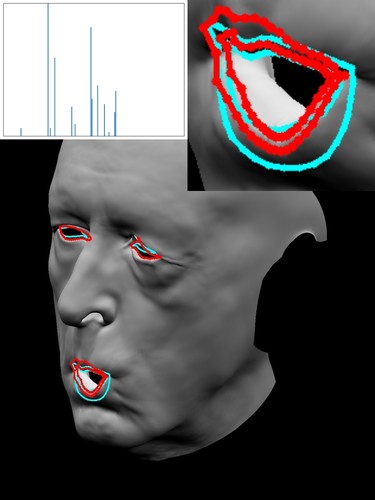}
    \caption{Our Approach}
\end{subfigure}
\begin{subfigure}[b]{0.17\linewidth}
    \includegraphics[width=\linewidth]{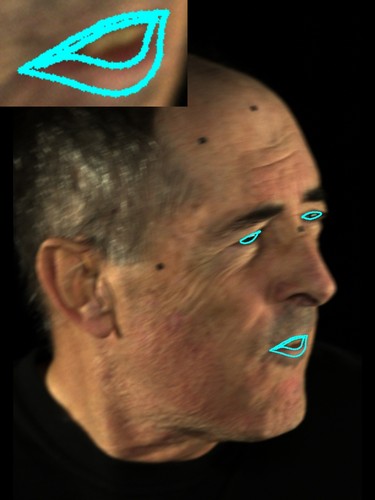}
    \includegraphics[width=\linewidth]{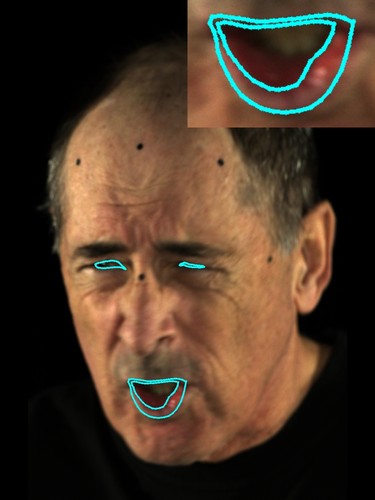}
    \includegraphics[width=\linewidth]{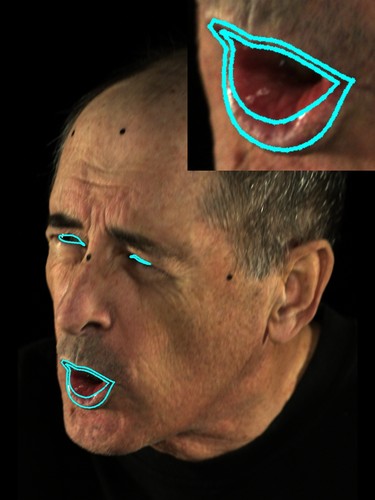}
    \includegraphics[width=\linewidth]{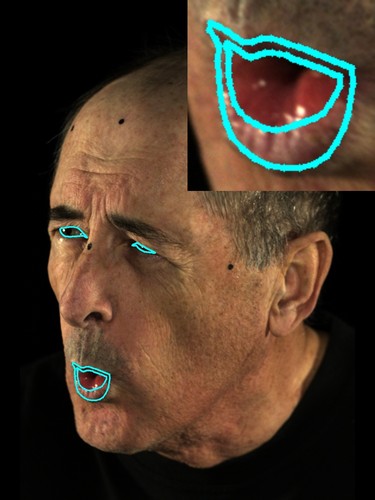}
    \caption{Target}
\end{subfigure}
\hfill
\caption{
Additional frames for the real image data example (Section 6.2).}
\label{fig:additional_plate_mesh_comparison}
\end{figure*}

Figure \ref{fig:additional_plate_mesh_comparison} shows additional frames for the real image data example (Section 6.2).  These additional frames help stress the consistency of our method in achieving its two goals.  First, our method produces facial expressions which reasonably approximate the corresponding targets.  Second, our method generally selects the sparset set of blendshape weights (with typically larger values) compared to other techniques, which means our results are easier to semantically interpret as particular human facial expressions.

\section{Extended Parameter Study}

\subsection{Column Choice}

We compare our approach for choosing the next coordinate descent column to using Gauss-Southwell and MBI.
For each approach, we linearize and solve with no thresholding for the relative decrease in L2 error, an upper limit of \num{10} unique coordinates used, and a fixed step size of \num{0.01}; in these examples, we remove the eye rotoscope curves from the energy function.
Even without solving for the eye rotoscope curves, MBI will still overfit and overdial mouth blendshapes (e.g.\ the two most dialed in shapes have magnitudes of \num{85.78} and \num{63.12}).
On the other hand, Gauss-Southwell and our approach maintain the face surface geometry and keep the blendshape weights within a reasonable range while maintaining the sparsity of the solution.
We note that with coordinate descent it is generally a matter of when, not if, the algorithm chooses a coordinate that will be overdialed; our examples demonstrate that MBI reaches those coordinates more quickly than Gauss-Southwell and our approach.
See Figure \ref{fig:appendix_column_index_comparison_no_eyes}.

\subsection{Step Size \& Convergence}

\begin{figure*}[t]
\centering
\begin{minipage}[t]{0.48\linewidth}
    \begin{subfigure}[t]{\dimexpr0.3\linewidth+10pt\relax}
        \makebox[10pt]{\raisebox{40pt}{\rotatebox[origin=c]{90}{First}}}%
        \includegraphics[width=\dimexpr\linewidth-10pt\relax]{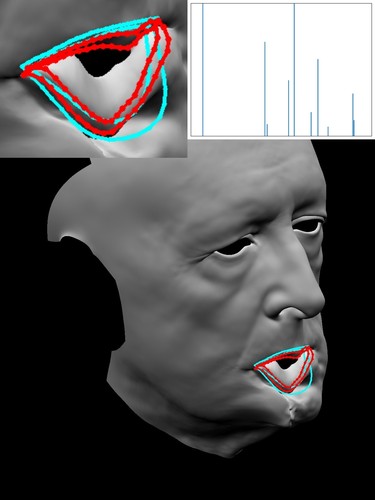}
        \makebox[10pt]{\raisebox{40pt}{\rotatebox[origin=c]{90}{Full}}}%
        \includegraphics[width=\dimexpr\linewidth-10pt\relax]{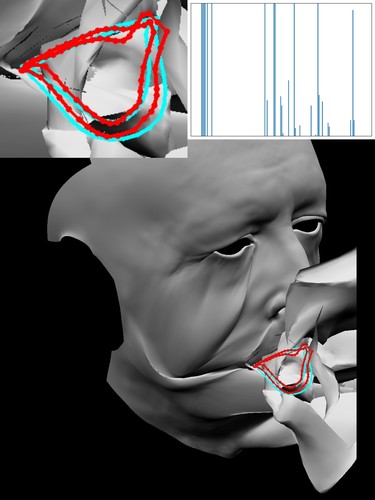}
        \caption{MBI}
    \end{subfigure}
    \begin{subfigure}[t]{0.3\linewidth}
        \includegraphics[width=\linewidth]{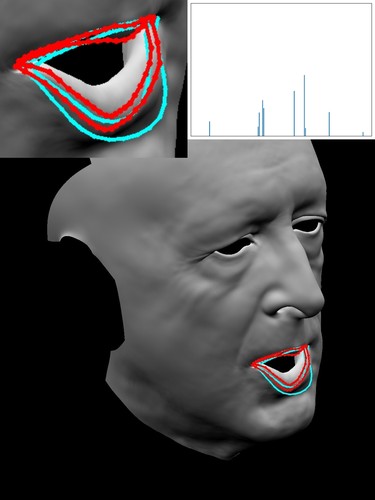}
        \includegraphics[width=\linewidth]{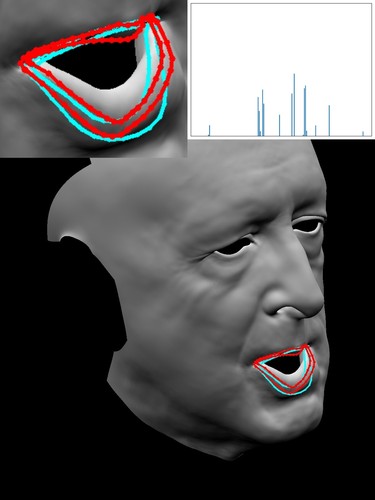}
        \caption{GS}
    \end{subfigure}
    \begin{subfigure}[t]{0.3\linewidth}
        \includegraphics[width=\linewidth]{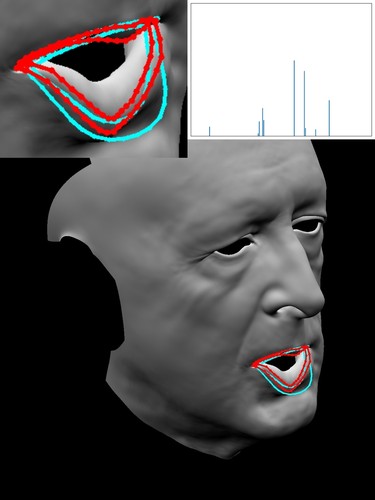}
        \includegraphics[width=\linewidth]{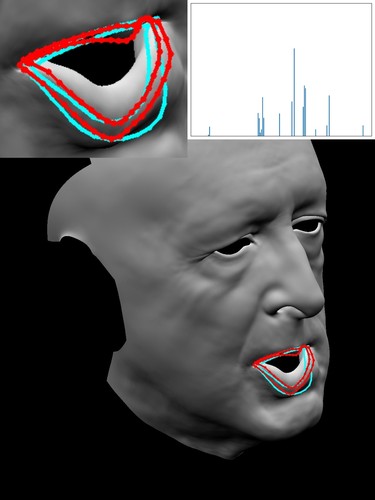}
        \caption{Ours}
    \end{subfigure}
    \hfill
    \caption{A comparison of the coordinates chosen by the MBI rule, the Gauss-Southwell (GS) rule, and our approach when solving without the eye rotoscope curves.
             The top row are the results after a single Gauss-Newton iteration, and the bottom row are the results after \num{10} Gauss-Newton iterations.}
    \label{fig:appendix_column_index_comparison_no_eyes}
\end{minipage}
\hfill
\begin{minipage}[t]{0.48\linewidth}
    \begin{subfigure}[t]{0.3\linewidth}
        \includegraphics[width=\linewidth]{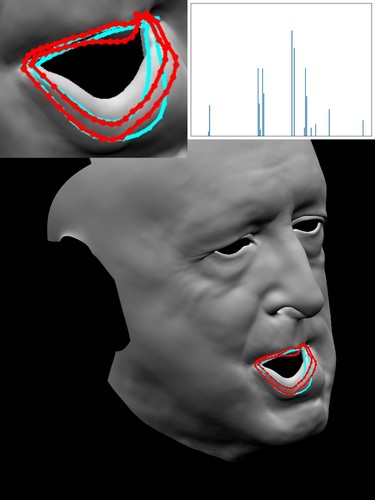}
        \caption{Greedy}
    \end{subfigure}\hfill
    \begin{subfigure}[t]{0.3\linewidth}
        \includegraphics[width=\linewidth]{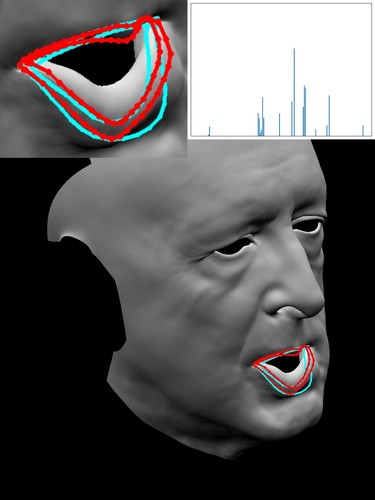}
        \caption{\num{0.01}}
    \end{subfigure}\hfill
    \begin{subfigure}[t]{0.3\linewidth}
        \includegraphics[width=\linewidth]{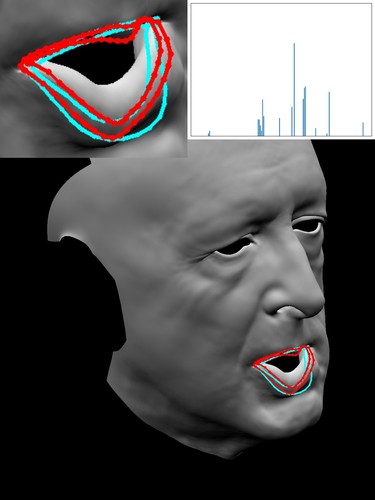}
        \caption{\num{0.02}}
    \end{subfigure}\hfill
    \begin{subfigure}[t]{0.3\linewidth}
        \includegraphics[width=\linewidth]{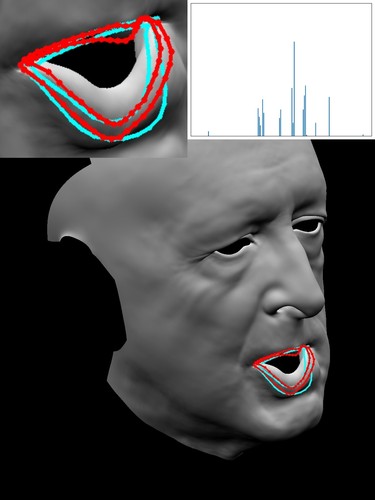}
        \caption{\num{0.1}}
    \end{subfigure}\hfill
    \begin{subfigure}[t]{0.3\linewidth}
        \includegraphics[width=\linewidth]{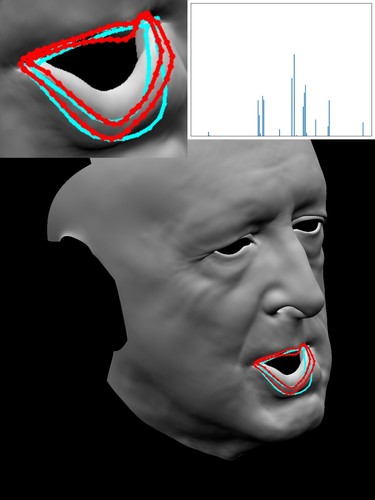}
        \caption{\num{0.5}}
    \end{subfigure}\hfill
    \begin{subfigure}[t]{0.3\linewidth}
        \includegraphics[width=\linewidth]{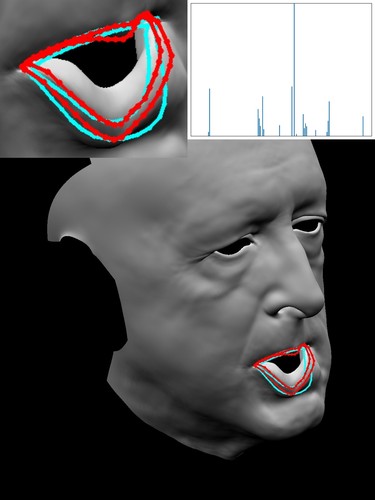}
        \caption{\num{1.0}}
    \end{subfigure}\hfill
    \caption{A comparison of the behavior of the geometry and the blendshape weights when using different step sizes when solving without the eye rotoscope curves.}
    \label{fig:appendix_step_parameter_comparison_no_eyes}
\end{minipage}
\hfill
\end{figure*}

As in Section 6.3, we compare several fixed step sizes of to the full, greedy step, i.e.\ $\alpha_i = r^T j_i / \| j_i \|_2^2$.
We again note that the fixed step sizes for each coordinate are clamped to not exceed the full, greedy step size.
Again without pruning, we run \num{10} Gauss-Newton iterations with no thresholding for the relative decrease in L2 error and an upper limit of \num{10} unique coordinates used; however, in these examples, we remove the eye rotoscope curves from the energy function.
In this case, the step size has no significant effect on the sparsity of the blendshape weights; however, as seen in Figure \ref{fig:appendix_step_parameter_comparison_no_eyes}, taking the greedy step versus taking a fixed size step results in a more unnaturally shaped mouth.
This would seem to indicate that always taking the greedy step will result in some overfitting.

\begin{figure*}[t]
\centering
    \begin{subfigure}[b]{0.48\linewidth}
        \includegraphics[width=\linewidth]{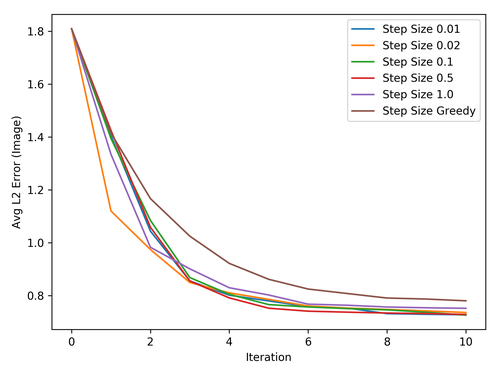}
        \caption{with eye rotoscope}
    \end{subfigure}
    \begin{subfigure}[b]{0.48\linewidth}
        \includegraphics[width=\linewidth]{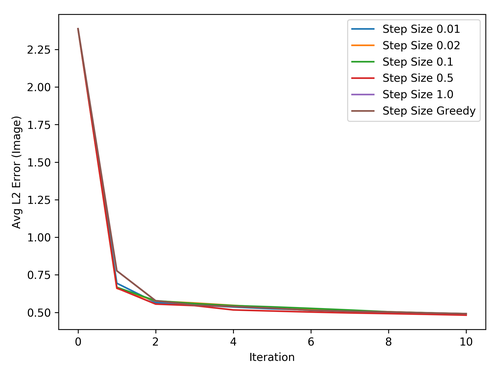}
        \caption{no eye rotoscope}
    \end{subfigure}
\hfill
\caption{A comparison of the average L2 errors plotted before every Gauss-Newton iteration when varying the step size parameter used to choose the next coordinate direction in our approach.
The brown lines plot the average L2 errors when using the Gauss-Southwell approach; notice how our approach allows for a faster reduction in error, particularly in the more difficult and realistic case where multiple sets of rotoscope curves are drawn.
}
\label{fig:appendix_isolate_step_parameter_errors}
\end{figure*}

\begin{figure*}[t]
\centering
\begin{subfigure}[b]{0.48\linewidth}
    \begin{subfigure}[b]{\dimexpr0.45\linewidth+10pt\relax}
        \makebox[10pt]{\raisebox{60pt}{\rotatebox[origin=c]{90}{Greedy}}}%
        \includegraphics[width=\dimexpr\linewidth-10pt\relax]{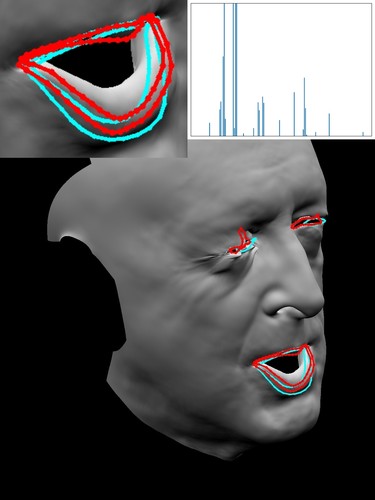}
        \makebox[10pt]{\raisebox{60pt}{\rotatebox[origin=c]{90}{0.01}}}%
        \includegraphics[width=\dimexpr\linewidth-10pt\relax]{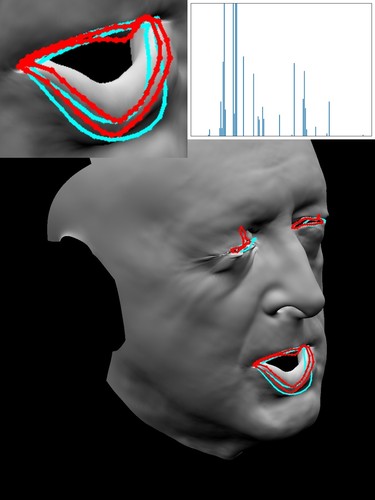}
        \makebox[10pt]{\raisebox{60pt}{\rotatebox[origin=c]{90}{0.02}}}%
        \includegraphics[width=\dimexpr\linewidth-10pt\relax]{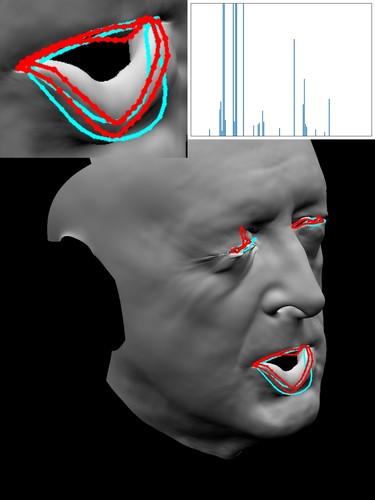}
        \hfill
        \caption*{with eye rotoscope}
    \end{subfigure}
    \begin{subfigure}[b]{0.45\linewidth}
        \includegraphics[width=\linewidth]{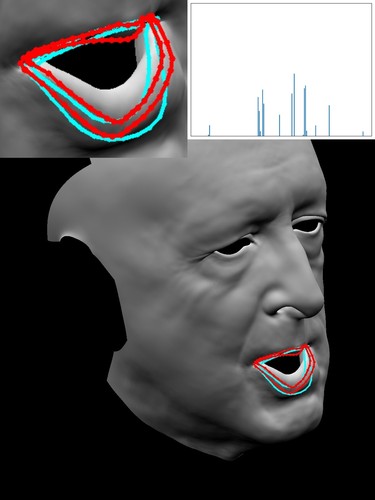}
        \includegraphics[width=\linewidth]{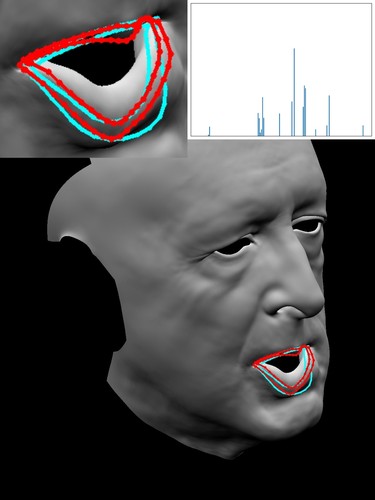}
        \includegraphics[width=\linewidth]{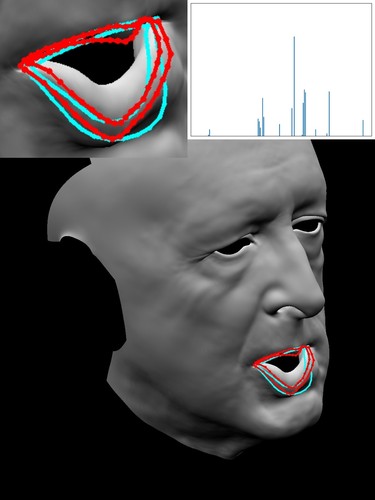}
        \hfill
        \caption*{no eye rotoscope}
    \end{subfigure}
    \hfill
\end{subfigure}
\begin{subfigure}[b]{0.48\linewidth}
    \begin{subfigure}[b]{\dimexpr0.45\linewidth+10pt\relax}
        \makebox[10pt]{\raisebox{60pt}{\rotatebox[origin=c]{90}{0.1}}}%
        \includegraphics[width=\dimexpr\linewidth-10pt\relax]{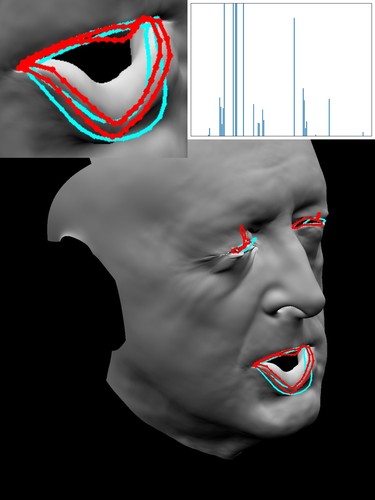}
        \makebox[10pt]{\raisebox{60pt}{\rotatebox[origin=c]{90}{0.5}}}%
        \includegraphics[width=\dimexpr\linewidth-10pt\relax]{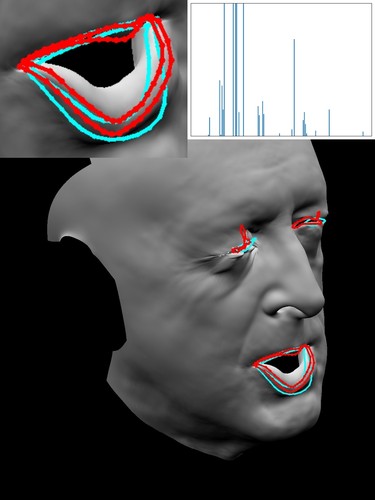}
        \makebox[10pt]{\raisebox{60pt}{\rotatebox[origin=c]{90}{1.0}}}%
        \includegraphics[width=\dimexpr\linewidth-10pt\relax]{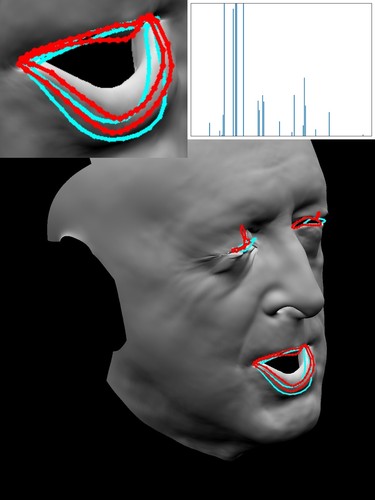}
        \hfill
        \caption*{with eye rotoscope}
    \end{subfigure}
    \begin{subfigure}[b]{0.45\linewidth}
        \includegraphics[width=\linewidth]{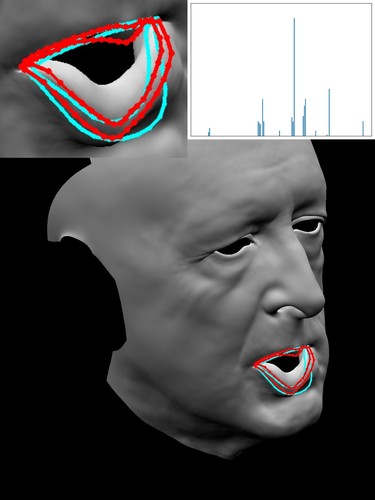}
        \includegraphics[width=\linewidth]{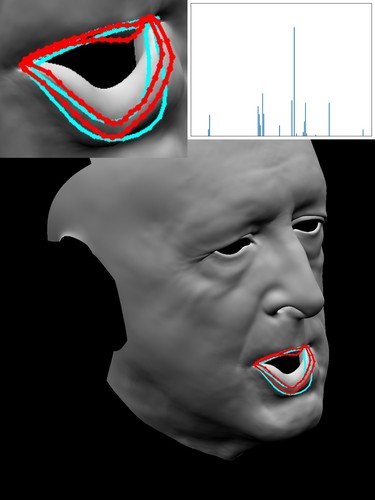}
        \includegraphics[width=\linewidth]{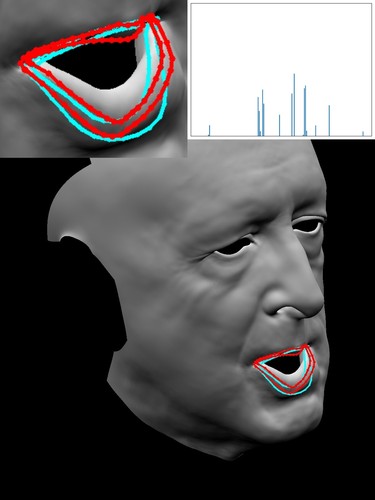}
        \hfill
        \caption*{no eye rotoscope}
    \end{subfigure}
    \hfill
\end{subfigure}
\hfill
\caption{A comparison of the geometry and blendshape results from varying the step size parameter used for choosing the next coordinate in our approach.}
\label{fig:appendix_isolate_step_parameter_comparison}
\end{figure*}

We also run tests with and without the eye rotoscope curves to isolate the effect the step size has on the column choice in our approach; while we vary the step size used to determine the next coordinate descent column, we fix the actual step size to have a magnitude of \num{0.01}.
Again without pruning, we run \num{10} Gauss-Newton iterations with no thresholding for the relative decrease in L2 error and an upper limit of \num{10} unique coordinates used.
We note that when the full, greedy step size is used to choose the next coordinate direction, our approach produces the same results, i.e.\ chooses the same coordinates and takes the same steps, as Gauss-Southwell.
As can be seen in Figure \ref{fig:appendix_isolate_step_parameter_comparison}, the final geometry and blendshape weights when varying the step size are similar to the results produced by Gauss-Southwell, i.e.\ using the full, greedy step.
However, our approach, as seen in Figure \ref{fig:appendix_isolate_step_parameter_errors} allows the error to decrease more in earlier Gauss-Newton iterations than when using Gauss-Southwell.
Therefore, it may be beneficial to use our approach with a fixed size step when only a few Gauss-Newton iterations are desired.

\section{Parameter Limits and Trust Region}

Using a coordinate descent solver for the linear problem at every Gauss-Newton iteration allows us to build the solution to $Ax = b$ incrementally without relying on a black box linear solver.
Any approach that aims to prevent the solution from going past a certain point (e.g.\ parameter limits and a trust region method) are thus trivial to implement.
Given min and max parameters $x^\text{min}_i$ and $x^\text{max}_i$, at every iteration of the coordinate descent solve, we can clamp each $\alpha_i$ such that $x^\text{min}_i \leq x_i + \delta x_i + \alpha_i \leq x^\text{max}_i$.
Additionally, given a trust region radius $\mu$, we can simply terminate the coordinate descent solve when/if $\|\delta x_S\|_2 > \mu$; this is similar in spirit to how the trust region parameter is used in the Dogleg method.
Furthermore, we can introduce limits on $\alpha_i$ to ensure that each step will not cause the solver to go beyond the trust region.

\end{document}